\documentclass[sigconf,nonacm]{acmart}

\usepackage[utf8]{inputenc} 
\usepackage[T1]{fontenc}    
\usepackage{hyperref}       
\usepackage{url}            
\usepackage{booktabs}       
\usepackage{amsfonts}       
\usepackage{nicefrac}       
\usepackage{microtype}      
\usepackage{xcolor}         
\usepackage{amsthm}

\usepackage{graphicx}
\usepackage{subfig}
\usepackage{caption}
\usepackage{float}
\usepackage{multirow}

\usepackage{array}

\usepackage{enumitem}
\setlist[itemize]{itemsep=0.1em, topsep=0.2em, parsep=0.1em, partopsep=0.1em}

\newtheorem{definition}{Definition}

\AtBeginDocument{%
  }

\setcopyright{acmlicensed}

\copyrightyear{2025}
\acmYear{2025}
\setcopyright{rightsretained}
\acmConference[SIGSPATIAL '25]{The 33rd International Conference on Advances in Geographic Information Systems}{November 3 - November 6, 2025}{Minneapolis, MN, USA}
\acmBooktitle{The 33rd Int. Conference on Advances in Geographic Information Systems (SIGSPATIAL '25), November 3 - November 6, 2025, Minneapolis, MN, USA}

\begin{document}
\title{Whose Truth? Pluralistic Geo-Alignment for (Agentic) AI}



\author{Krzysztof Janowicz}
\email{krzysztof.janowicz@univie.ac.at}
\affiliation{%
  \institution{University of Vienna}
  \city{Vienna}
  \country{Austria}
}


\author{Zilong Liu}
\email{zilong.liu@univie.ac.at}
\affiliation{%
  \institution{University of Vienna}
  \city{Vienna}
  \country{Austria}
}

\author{Gengchen Mai}
\email{gengchen.mai@austin.utexas.edu}
\affiliation{%
  \institution{University of Texas at Austin}
  \city{Austin}
  \country{USA}
}

\author{Zhangyu Wang}
\email{zhangyu.wang@maine.edu}
\affiliation{%
  \institution{University of Maine}
  \city{Orono}
  \state{Maine}
  \country{USA}
}

\author{Ivan Majic}
\email{ivan.majic@univie.ac.at}
\affiliation{%
  \institution{University of Vienna}
  \city{Vienna}
  \country{Austria}
}

\author{Alexandra Fortacz}
\email{alexandra.fortacz@univie.ac.at}
\affiliation{%
  \institution{University of Vienna}
  \city{Vienna}
  \country{Austria}
}

\author{Grant McKenzie}
\email{grant.mckenzie@mcgill.ca}
\affiliation{%
  \institution{McGill University}
  \city{Montr\'eal}
  \country{Canada}
}

\author{Song Gao}
\email{song.gao@wisc.edu}
\affiliation{%
  \institution{University of Wisconsin}
  \city{Madison, WI}
  \country{USA}
}

\renewcommand{\shortauthors}{Janowicz et al.}

\ccsdesc[500]{Social and professional topics~Geographic characteristics}
\ccsdesc[500]{Computing methodologies~Philosophical/theoretical foundations of artificial intelligence}
\ccsdesc[500]{Applied computing}

\keywords{GeoAI, Geo-Alignment, Pluralistic Alignment, AI Ethics, Bias} 

\sloppy

\begin{abstract}
    AI alignment describes the challenge of ensuring (future) AI systems behave in accordance with societal norms, values, and goals. Alignment is now central to research on foundation models and AI agents. Most recent work focuses on methods to prevent potentially harmful biases, account for social inequalities, improve AI safety, and enhance explainability. Notably, the debiasing `corrections' applied to various stages of AI/ML workflows may lead to outcomes that diverge strongly from current statistical realities on the ground. For instance, text-to-image models may depict a balanced gender ratio of company leadership, despite existing imbalances. However, an often overlooked dimension is the \textit{geographic variability} of alignment. What is considered appropriate, truthful, or legal can vary greatly between regions due to cultural differences, political realities, or legislation. Hence, some model outputs align without further knowledge of the user's geospatial context, while others are highly sensitive to it. Put differently, whether these outputs align varies geographically. E.g., statements about Kashmir cannot be generated without understanding the user's origin and current location. From a common-sense perspective, this problem is hardly new. In fact, Google Maps will render different administrative borders based on the user's location. Interestingly, in both knowledge representation and representation learning, spatiotemporal context, e.g., due to the monotonic nature of reasoning, remains a major challenge. Until very recently, these were largely theoretical problems. What is truly novel is the scale and level of automation at which AI systems now mediate knowledge, express opinions, and represent \textit{reality} to millions of users across borders, often with little transparency or oversight regarding how context is handled. With agentic AI on the horizon, the urgency for pluralistic, geographically aware alignment, rather than one-size-fits-all solutions, is growing. Here, we motivate and formalize the \textbf{vision of geo-alignment}, outline how it goes beyond pluralistic alignment by offering learnable spatially explicit patterns, and suggest concrete avenues for future research. 
\end{abstract}

\maketitle

\section{Introducing Geo-Alignment}

In a nutshell, \textit{alignment} asks the question of \textit{how to ensure AI systems behave according to societal norms, values, and goals, even though those goals may themselves be complex, contested, context-dependent, or even conflicting}. A system is \textit{aligned} if its actions advance these norms, values, and goals, and \textit{misaligned} if it acts otherwise \cite{wiener1960some,russell2019human}. The challenge of \textit{super-alignment} goes one step further by focusing on (potential) future super-intelligence. This not only changes the urgency, but also the scope and methods, as such systems may reason beyond our abilities, pursue goals we cannot comprehend, and act at speeds that render post-hoc oversight impossible \cite{guanDeliberativeAlignmentReasoning2025}.

Unfortunately, however, established research and practice often glosses over the fact that the goals of individuals, groups, and societies do not always converge. Put differently, \textbf{existing work risks confusing inter-human variation with noise} \cite{sorensen2024roadmap}. This was acceptable as long as AI alignment was largely a philosophical endeavor, but how is an AI agent supposed to align if we humans cannot? Hence, AI research has begun to account for \textit{value pluralism} \cite{gabriel2020artificial} and \textit{pluralistic alignment} \cite{sorensen2024roadmap}. For example, the authors distinguish between \textbf{(1)} \textit{overton pluralism}, where all potentially `reasonable' answers are returned; \textbf{(2)} \textit{steerable pluralism}, where the model outputs are guided towards a given perspective (called an attribute); and \textbf{(3)} \textit{distributional pluralism}, where outputs are drawn in accordance with those of a desired reference population. 

Geography, as defined by the interplay of its spatial and temporal structure through dependence and heterogeneity, and the societal layer arising on top, has not been prominently featured in alignment research so far. This is a problem because it risks AI systems producing outputs that are potentially locally inappropriate, culturally insensitive, or even illegal. For example, a system may recommend medication available over-the-counter in the US but strictly regulated in other regions or for other contexts where it might be classified as doping or be an illegal substance altogether. In this paper, we therefore argue for a \textbf{geographic perspective on alignment}, particularly in light of the current rise of agentic, autonomous GeoAI systems (GeoMachina) \cite{janowicz2020geoai, li2025giscience}. 

Such a \textit{geo-alignment} goes beyond pluralistic approaches in also investigating (and, ultimately, learning) the spatial structure underlying the perspectives to which AI systems should align. This addresses a core research gap in the literature, namely, \textbf{how to derive `reasonable' perspectives without overburdening (human) expert annotators}. While we are refraining from making an overly strong \textit{geographic determinism} argument here, spatial structure enables prediction, which current approaches struggle with. We believe that lessons learned from studying the role of spatial dependence (measures) on the scaling laws of GeoAI models \cite{wang2024probing} can therefore inform work on AI alignment. Simply put, similar (e.g., nearby) regions may have similar alignment needs, and this relation will decay with distance -- not necessarily a Euclidean norm due to so-called \textit{place-based effects} -- in a predictable and learnable manner. This opens up an entirely \textbf{new avenue} for approaching alignment in the GeoAI literature, where it has received little attention to date. 

\textbf{The contributions of this vision paper are as follows:
}
\begin{itemize}
    \item We introduce the problem of AI alignment to the GeoAI community, where it has received marginal attention so far.
    \item Ensuring GeoAI is not a one-way street, we discuss how geo-alignment may benefit the broader AI community.
    \item We outline how geo-alignment could be formalized and measured using context and spatial structure. Unlike other dimensions of pluralism, spatial structure is predictable/learnable.
    \item We present vignettes for future geo-alignment research. 
\end{itemize}

In the following, we briefly review prior work and draw connections to AI alignment, thereby establishing links that have not been previously made by the GeoAI community, e.g., between geo-privacy and AI agent memory. Then, we outline a formalization for geo-alignment and provide selected vignettes for future research. 

\section{Relating Prior Work to Alignment}
Here, we briefly review GeoAI work on bias, trust, and ethics more generally, and link them back to AI alignment, i.e., \textit{the problem of control} \cite{russell2019human}.

\vspace{-0.15cm}
\paragraph{\textbf{Bias and the Geo-Diversity of Models}}
Research on AI bias studies how to detect, quantify, and correct imbalances in training data, learned representations, and model outputs in order to produce more equitable or representative results \cite{mehrabi2021survey}. Among all kinds of biases, geographic bias has quickly become an important yet less-studied research topic. A few pioneering studies demonstrated that many widely used datasets exhibit strong geographic bias. For instance, \citet{shankar2017no} revealed that only about 3\% of all images in their ImageNet sample were from China and India combined, while 45\% were from the US alone. Such bias in training data may lead to disparities in model performance across geographic regions \cite{shankar2017no,ramaswamy2023geode,liu2022geoparsing}. Due to geographic bias in the training data, downstream models also exhibit geographic bias. \citet{faisal2023geographic} demonstrated that the learned representations of language models show strong \textit{geopolitical favouritism} wrt. certain regions (e.g., countries with higher GDP). \citet{manvi2024large} demonstrated that large language models (LLMs) exhibit bias against regions with lower socioeconomic conditions on sensitive, subjective topics. Both \citet{wu2024torchspatial} and \citet{wang2025geobs} demonstrated that vision and multimodal foundation models exhibit bias towards developed countries. They developed scores to further quantify such bias.

Bias and alignment intersect where the lack of representation in model inputs can cause systems to discriminate, e.g., by advancing stereotypes or marginalizing populations or regions through underrepresentation in model outputs. From the perspective of AI development and deployment, such problems can be corrected by implementing debiasing methods. Unfortunately, however, these methods introduce new problems of their own. Who decides what, how, and to what degree to \textit{debias}, especially if existing training data collides with human values for the future of our societies? Altering the distribution of genders in the model's output may produce a wanted effect of promoting gender equality, but non-selective debiasing may also cause some unwanted outputs that are factually and morally incorrect. As a notable example, Google's Gemini depicted black and female Nazi soldiers as a result of overcorrective attempts to increase ethnic and gender representation\footnote{For example: \url{https://www.nytimes.com/2024/02/22/technology/google-gemini-german-uniforms.html}}. 

Interestingly, alignment may also introduce new biases in model outputs. Recently, \citet{liu2025operationalizing} proposed that the most observed kind of geographic bias in both model inputs and outputs is \textit{a lack of geographic diversity}. They further discovered that as AI chatbots have advanced, the geo-diversity of their output has also decreased. A similar finding has recently been reported by~\citet{murthy2024one}, suggesting that this could be a direct, unfortunate consequence of efforts to reduce the hallucination rate of the underlying models.

\vspace{-0.15cm}
\paragraph{\textbf{Trust and Explainability}} 
Bias within training data leads to questions of trust. Trust, however, extends far beyond just trust in training data, but also to trust in the models themselves, including how the models were developed, who decided on (political) debiasing measures, and whose cultural norms act as reference for alignment. As it turns out, these aspects are also geographic. As a \textit{human-in-the-loop} strategy, Reinforcement Learning from Human Feedback (RLHF) is itself sensitive to geographic regions, for example, through linguistic and cultural differences. Research on truly multi-lingual alignment via RLHF is still in its infancy \cite{lai2023okapi}. This is a problem, as numerous existing studies have shown a diversity crisis in AI development~\cite{west2019discriminating} with many models developed by (white) men from western cultures and the majority of data labeling and curation being done by individuals from the Global South~\cite{casilli2025global}. Yet, still very little GeoAI research has explicitly investigated how a lack of cultural and geographic model diversity impacts user trust.

The relationship between misalignment and trust has so far been underexplored. While numerous authors have called for a deeper investigation into the value alignment between ``humans and machines''~\cite{russell2019human}, to date, little effort has attempted to quantify the (mis)alignment, let alone assess contributing factors. Here, again, we see a clear direction for future research formalizing regional and cultural variation in value alignment and trust.

\vspace{-0.15cm}
\paragraph{\textbf{Ethics of GeoAI}} 
While work on bias and trust are among the common themes studied in the ethics of GeoAI, the field is substantially broader. It also includes algorithmic fairness, explainability, privacy, and sustainability \cite{janowicz2023philosophical}. Many of these could benefit from being linked to alignment research, although this has not yet occurred. For instance, research on geo-privacy raises alignment questions about how systems should balance model utility with the protection of sensitive, locational information \cite{rao2023building,rao2023cats}. Intuitively, increasing contextual information improves an agent's ability to align with a user's goals, \textit{but at a cost that is currently not well understood}. A very recent example is user concerns about ChatGPT's memory function and OpenAI's response to mitigate these potential negative privacy effects\footnote{\url{https://help.openai.com/en/articles/8590148-memory-faq}}. Put differently, geo-privacy research can not only inform the alignment debate but, conversely, alignment raises its own novel geo-privacy problems.

\section{Towards a Formal Theory of Geo-Alignment}
Here we provide an intuition about how geo-alignment could be formally defined and measured.

Let: 
\begin{itemize}
    \item $q\in Q$ be a user's query;
    \item $o \in O$ be an AI system's output;
    \item $g \in G$ be the geographic context (e.g., a $<s,t>$--tuple representing space $s$ and time $t$);
    \item $D(\cdot\,,\cdot)$ be a dissimilarity or divergence measure;
    \item $L(o\,|\,q,g)$ be the conditional probability distribution of the (hypothetical) locally appropriate outputs given a query and geographic context (e.g., a region); 
    \item $S(o\,|\,q,g)$ be the conditional probability distribution of outputs of AI system $s$ at $g$;
    \item $\epsilon$ be a scalar for the upper bound of acceptable dissimilarity between the compared distributions.
\end{itemize}


\begin{definition}
 An AI system is said to be \textbf{geo-aligned} if for all queries $q\in Q$ and geographic contexts $g \in G$, the distribution of the system's output $S(o\,|\,q,g)$ resembles the locally appropriate distribution of outputs $L(o\mid q,g)$ 
 within a certain tolerance.
\end{definition}

 More formally: 
 {\large \begin{equation}
    \forall q \in Q, \forall g \in G{:\,} D(L(\cdot|q,g), S(\cdot|q,g)) < \epsilon
 \end{equation}}

For model training, it can be more useful to measure the `degree' of geo-alignment in evaluations. In such cases, we need a penalty function $f$ which reflects how much $S$ deviates from $L$ on average. For example, $f$ can be as simple as the mathematical expectation of $D$ over all query/output pairs, with the caveat that this penalty will overlook low-probability cases.

 Now, for any given query and geographic context, $L(o\,|\,q,g)$ may be difficult to estimate as \textbf{(1)} the boundaries of regions may be vague\footnote{In fact, theoretically speaking, there may be infinitely many of them -- an issue that can be addressed by learning the spatial structure and/or using a hierarchical discrete global grids such as S2 as will be discussed later.} or data-poor, \textbf{(2)} queries may be complex or occur rarely, and \textbf{(3)} an acceptable $\epsilon$ may vary across $q$, $g$, and use case. The latter two are well-known (sparsity) issues in the general alignment literature, and, thus, our focus will be on (1).

Unfortunately, most interesting (geo-)alignment problems are less clear-cut than, say, legal variations in drinking age. Instead, they revolve around nuanced but impactful differences in cultural norms, local attitudes, and societal expectations. For instance, while an LLM and Retrieval Augmented Generation (RAG) system can provide accurate replies on drinking age, it is unlikely to pick up on the different trade-offs (many) Europeans and Americans make when balancing privacy concerns with technological convenience without explicit context (e.g., cross-session agent memory). 

To provide a concrete example, the status of pseudoephedrine varies significantly across regions and periods. It has been an over-the-counter drug in some countries, prescription-only or even illegal in others, and has been classified as a monitored or even banned performance-enhancing substance. As of June 2025, several AI systems return a US-centric answer only, while a few others either rely on the user's location or provide selected national variations. No system identified the potential problems related to sports.
\begin{equation}
    D_{\text{KL}}(P\parallel Q)=\sum _{x\in {\mathcal {X}}}P(x)\,\log {\frac {P(x)}{Q(x)}}
    \label{KL}
\end{equation}
As a simplified example, let $D(\cdot\,,\cdot)$ be Kullback-Leibler (KL) divergence (Eq. \ref{KL}) and let the locally appropriate distribution $L(o\,|\,q,g)$ for "\textit{Can I buy pseudoephedrine over the counter?}" ($q$) and the United States ($g$) be $\{0.8,0.15, 0.05 \}$ for outputs ($o$) "With an ID.", "Requires prescription." and "Careful with doping.", respectively. Now, if a system's outputs $S_1(o\,|\,q,g)$ are $\{0.998,0.001,0.001\}$, $D_{\text{KL}}$ would be 0.771, while for a system $S_2$ returning $\{0.7,0.2,0.1\}$ $D_{\text{KL}}$ would be 0.029 if using the natural logarithm. Simply put, in this hypothetical example, $S_1$ is less geo-aligned as it only returns the "With an ID." answer despite regulations varying by state. While $S_2$ overestimates the number of people interested in an answer merely relevant to competitive sports, its overall geo-alignment is better.

Intuitively, one may argue that Overton pluralism already captures this case: an AI system should simply return all `reasonable' cases. However, this is exactly why we picked the pseudoephedrine example. In fact, the regulations not only differ across use cases and nations but also across multiple US states that regulate how it can be sold, how much of it can be sold in a single transaction, and during which period. Even more, these regulations are still changing.\footnote{For instance, since 2022, Oregon and Mississippi no longer require a prescription.} Thus, returning pages of legal text to the user is not a solution. After all, the \textit{synthesis} modern AI bots provide is what sets them apart from the WWW. However, this synthesis requires spatial and temporal context to understand what matters \textit{where} and \textit{when}. This is exactly what our geo-alignment vision is all about.





\section{Directions for Research}
In the following, we present brief vignettes that highlight novel research challenges related to geo-alignment and situate them within the broader SIGSPATIAL and GeoAI research community.

\vspace{-0.15cm}
\paragraph{\textbf{Vignette 1: Geographic Alignment Benchmarks}} 

Our proposed formalization is grounded in the observation that core aspects (e.g., societal norms) of geo-alignment are inherently collective -- rather than individual -- as they are shared across populations, thereby making geo-alignment a pluralistic endeavor. Geographers typically analyze collective phenomena through the lens of regions, making regional context, e.g., states, a natural unit of analysis for geo-alignment. Therefore, one priority would be to develop benchmarks with a stronger focus on societal norms, values, and goals, in order to later evaluate a system's alignment with local knowledge and customs. Such efforts not only reflect regional variability, but also offer a means to investigate how existing AI models potentially misrepresent local populations and their sense of place~\citep{tuan1979space}. Recent work already suggests that system outputs show strong geographic \textit{defaults} \cite{liu2025operationalizing}, which is highly problematic but hardly surprising given findings by \citet{ballatore2017digital} about (a lack of) \textit{localness} in search results. In terms of GeoAI research, developing such benchmarks would also build on recent work in the area of spatial representation learning \cite{mai2022review,mai2024srl}, which considers (general) geo-bias, but not yet alignment \cite{wu2024torchspatial}.

Geo-alignment also extends beyond classical geographic feature types — such as \texttt{forest} or \texttt{building} — which have received considerable attention in both the spatially explicit machine learning and the geo-semantics literature \cite{bennett2010spatial} over the past decades. Until very recently, most knowledge representation work considered concepts such as \texttt{wedding dress} \cite{yin2022geomlama} detachable from geographic space. However, they are becoming increasingly central to AI (alignment and bias) evaluation.\footnote{Most text-to-image models over-proportionally (or even exclusively) create white wedding dresses, while several cultures use the colors red or yellow.} Finally, many (geo-)alignment problems are rooted in territorial disputes (see \citet{li2024land}). 

How can we maximize the benefits of the benchmarks we construct? One approach is to leverage them to study how users from regions worldwide interact with (agentic) AI systems. For example, recent work by \citet{qadri2023ai} demonstrates how community-centered studies can reveal value misalignments in text-to-image models within the South Asian context.

\vspace{-0.15cm}
\paragraph{\textbf{Vignette 2: Neurosymbolic Approaches to Geo-Alignment}}

From a future research perspective, it is worth noting that while classical geo-semantics (i.e., symbolic/declarative) approaches can represent multiple, even contradictory, perspectives, they struggle to reason under such scenarios and with noise and missing data in general \cite{nickel2015review}. Representative learning approaches can handle such noise well, but struggle with implicit spatial and temporal context, and explainability. Additionally, their latent reasoning capabilities remain very limited. Combining the best of both worlds is where novel work on neuro-symbolic GeoAI comes into play \cite{mai2022symbolic}.


In a bid to improve the safety of their models, OpenAI recently introduced \emph{deliberative alignment} - a training approach that guides LLMs to explicitly reason through safety specifications before producing an answer~\cite{guanDeliberativeAlignmentReasoning2025}. Because the \textit{declarative} safety specifications are too large to fit into the model's context window at run time, they make the model learn their sub-symbolic representations during training instead. This also teaches the model to recognize when a policy might be relevant and then reason over the specifications to produce a compliant (or aligned) output. Although this approach is non-geographic so far, it provides a promising starting point for combining symbolic and neural approaches for geo-alignment, while also benefiting from geo-alignment, which may enable local variations in safety alignment.

The first requirement for achieving deliberative geo-alignment would be a geographical equivalent to declarative safety specifications. Safety alignment is often simpler than geo-alignment since there are only two possible cases. The output is only generated for safe cases, while the unsafe cases trigger a templated response, i.e., unsafe queries are refused. Geo-alignment would require a comprehensive spatially-explicit declarative specification that has a graded structure, as there may be many regions and periods to be learned during training time. Approaching the problem from a RAG perspective is a new avenue and supports on-the-fly geo-alignment (for changing contexts). Geo-knowledge graphs \cite{zhu2025knowwheregraph} may be suitable for this purpose and are being used as RAG providers by many LLM bots already. Such graphs rely on declarative knowledge representation and thus can model spatial and temporal scopes. 

The second requirement would be to teach the model to use the sub-symbolic representations to (1) recognize when a query requires geo-alignment (e.g., \texttt{"What is the speed of light?"} does not), and (2) to perform geo-alignment during reasoning in order to produce geo-aligned outputs. Such training would require a 
differentiable measure to guide the geo-alignment process.

\vspace{-0.2cm}
\paragraph{\textbf{Vignette 3: Learning from Spatial Structure}}
Early work reveals that unsupervised style transfer across language corpora can be achieved by aligning their latent Platonic representations \cite{shen2017style, huh2024platonic}. This is inspiring for geo-alignment because there are natural spatial structures (e.g, spatial autocorrelation and topology) which can be used. 
More specifically, we can align AI models by explicitly applying the constraints that 
ensure outputs, such as images or text, to demonstrate similar spatial structures.
Also, aligning spatial structures, especially in the latent representation space, allows spatial reasoning (e.g., hierarchical chains-of-thoughts~\cite{wei2022chain}). Simply put, a model can learn the spatial structure \cite{mai2022review} first and then use this to reason about geographic context similarity to overcome the sparseness issue discussed above. After all, nearby regions are more likely to have similar regulations and customs. Similarly, we can use a calculus over a hierarchical discrete global grid (e.g., S2) to
infer context. E.g., regulations for a region also hold for its subregions.


\section{Conclusions}
In this paper, we introduced the vision of geo-alignment, arguing that spatial structure is not just another dimension of pluralism in the alignment literature (which, by itself, is a very novel topic) but a source of learnable patterns that can be leveraged for alignment and evaluation. Such geo-alignment will be key to agentic AI systems as they will act in real geographic space. We proposed a formalization for geo-alignment, provided vignettes to outline future research directions, and established \textit{bidirectional} connections between GeoAI literature and current AI alignment research to demonstrate how they benefit from one another.




\bibliographystyle{ACM-Reference-Format}
\bibliography{reference}

\end{document}